\documentclass[10pt,twocolumn,letterpaper]{article}

\usepackage[pagenumbers]{cvpr} 

\usepackage{multirow}
\usepackage{colortbl}
\usepackage{textcomp}
\usepackage{rotating}
\usepackage{placeins}
\newcommand{\sysname}{3D-LLaVA }

\definecolor{tabblue}{rgb}{0.12,0.49,0.85}

\definecolor{cvprblue}{rgb}{0.21,0.49,0.74}
\usepackage[pagebackref,breaklinks,colorlinks,allcolors=cvprblue]{hyperref}

\def\paperID{2992} 
\def\confName{CVPR}
\def\confYear{2025}

\title{3D-LLaVA: Towards Generalist 3D LMMs with Omni Superpoint Transformer}

\author{Jiajun Deng$^1$, Tianyu He$^2$, Li Jiang$^3$, Tianyu Wang$^4$, Feras Dayoub$^1$, Ian Reid$^{1,4}$\\
$^1$ Australian Institute for Machine Learning, The University of Adelaide \\
$^2$ Microsoft Research \quad\quad $^3$ The Chinese University of Hong Kong, Shenzhen \\
$^4$ Mohamed bin Zayed University of AI\\
\url{https://github.com/djiajunustc/3D-LLaVA}.
}

\begin{document}
\maketitle
\begin{abstract}
Current 3D Large Multimodal Models (3D LMMs) have shown tremendous potential in 3D-vision-based dialogue and reasoning. However, how to further enhance 3D LMMs to achieve fine-grained scene understanding and facilitate flexible human-agent interaction remains a challenging problem. In this work, we introduce \textbf{3D-LLaVA}, a simple yet highly powerful 3D LMM designed to act as an intelligent assistant in comprehending, reasoning, and interacting with the 3D world. Unlike existing top-performing methods that rely on complicated pipelines—such as offline multi-view feature extraction or additional task-specific heads—3D-LLaVA adopts a minimalist design with integrated architecture and only takes point clouds as input. At the core of 3D-LLaVA is a new Omni Superpoint Transformer (OST), which integrates three functionalities: (1) a \textbf{visual feature selector} that converts and selects visual tokens, (2) a \textbf{visual prompt encoder} that embeds interactive visual prompts into the visual token space, and (3) a \textbf{referring mask decoder} that produces 3D masks based on text description. This versatile OST is empowered by the hybrid pretraining to obtain perception priors and leveraged as the visual connector that bridges the 3D data to the LLM. After performing unified instruction tuning, our 3D-LLaVA reports impressive results on various benchmarks.

\end{abstract}    
\section{Introduction}
\label{sec:intro}

\begin{figure}
    \centering
    \includegraphics[width=0.98\linewidth]{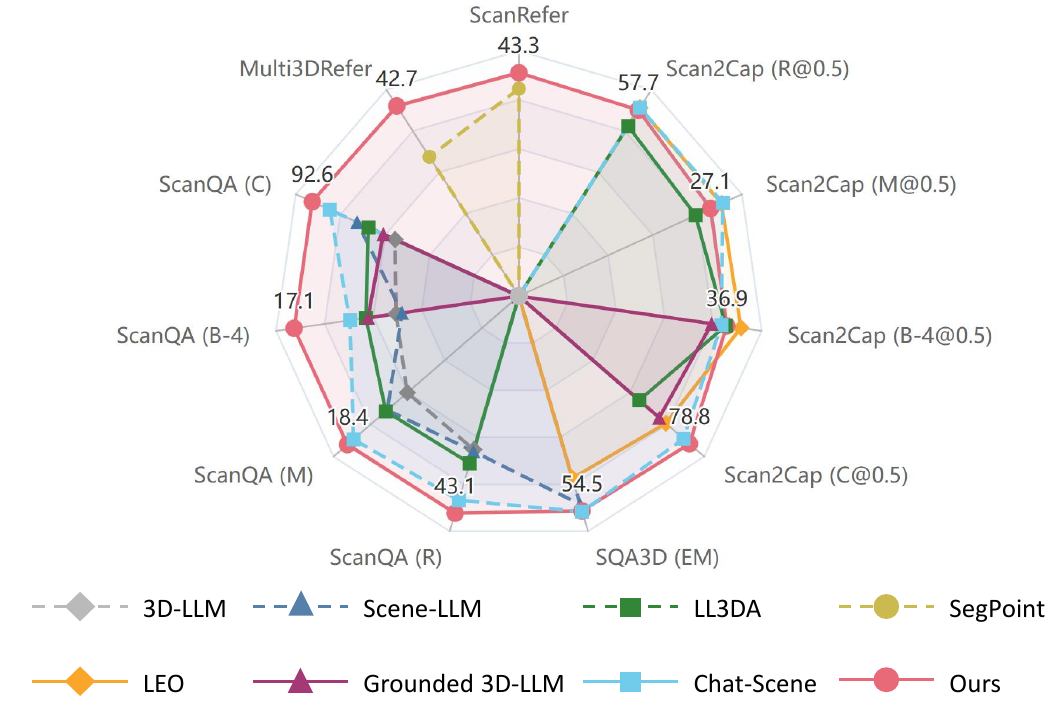}
    \vspace{-0.2cm}
    \caption{An intuitive comparison between 3D-LLaVA and other SoTA 3D LMMs (The performance of LEO on ScanQA is omitted here since its setting is different). Our 3D-LLaVA achieves the best results among the competitors on most of the benchmarks. }
    \label{fig:intro}
    \vspace{-0.4cm}
\end{figure}

Recent advancements in Large Language Models (LLMs)~\cite{brown2020gpt3,radford2019gpt2,zhu2023minigpt4,touvron2023llama,vicuna2023} have reshaped the paradigm of artificial intelligence, positioning language as a universal interface for general-purpose reasoning and interaction. Building on this progress, 2D Large Multi-modal Models (LMMs)~\cite{liu2023llava,liu2024improved,li2023blip2,instructblip,alayrac2022flamingo} have emerged, integrating images and texts to support a wide range of vision-language tasks. In a further step to extend these capabilities to 3D, 3D LMMs~\cite{hong20233d-llm,chen2024ll3da} have huge potential to unlock a series of real-world applications, such as autonomous vehicles, household robots, and augmented reality, where robust reasoning, precise 3D scene comprehension, and seamless human-agent interaction are of great significance.

It is a non-trivial problem to empower 3D LMMs with these desired properties. Despite notable progress achieved with the 3D vision and language community towards 3D dialogue and reasoning, these 3D LMMs still rely on extra prompt encoder~\cite{liu2023llava} or offline region proposal generation and feature extraction~\cite{huang2024chat,hong20233d-llm} to enable interacting with both visual and textual prompts. Such extra modules and offline preprocessing result in a complex pipeline, which complicates deployment and limits accessibility.

Furthermore, an effective 3D vision and language assistant should extend beyond simply generating text output; it should also be capable of grounding open-ended language expressions within a 3D scene and accurately segmenting the corresponding 3D masks. However, current referring-based 3D point cloud segmentation methods typically align and fuse text embeddings into a specialized segmentation model without LLMs. An exception is SegPoint~\cite{he2025segpoint}, which utilizes the reasoning capabilities of LLMs to improve referring 3D point segmentation. Nonetheless, it still depends on additional modules to achieve precise segmentation, and has not demonstrated its effectiveness in other 3D vision-language tasks such as VQA and captioning.

To overcome these limitations, we present \textbf{3D-LLaVA}, a generalist 3D LMM that streamlines the pipeline while maintaining strong performance across diverse 3D tasks. 
In contrast to the prior works that assemble multiple models or extract features in an offline manner, 3D LLaVA bridges interactive 3D vision dialogue and point-level 3D scene comprehension in an integrated and shared architecture, eliminating the need for auxiliary modules and complicated steps. Particularly, as compared in Figure 1, the most distinguishing part of 3D-LLaVA is the novel visual connector, namely Omni Superpoint Transformer (OST). What distinguishes it is how we use it as a shared module for multiple purposes and how we pretrain it with the 3D scene encoder.

Specifically, existing 3D LMMs generally follow the trend of the 2D domain to leverage an MLP projector~\cite{liu2023llava} or Q-Former~\cite{li2023blip2} as the visual connector, both of which are single-function modules designed to transform vision features into token embeddings aligned with the language semantic space.
On the contrary, OST is a versatile module built on superpoint representation that plays multiple roles in our 3D-LLaVA. Specifically, in addition to feature enhancement and projection, OST has the following functions:
(1)\underline{Visual Feature Selector.} OST selectively retains visual tokens, distinguishing between foreground and background superpoints. This helps highlight the informative part of the complex 3D scene and manage computational ovehead by reducing the number of tokens to be further processed by the LLM.
(2) \underline{Visual Prompt Encoder.} 3D-LLaVA does not involve an additional visual prompt encoder. When the user interacts with 3D-LLaVA with a visual prompt (such as a clicking point, a box, or a mask), OST plays the role of a visual prompt encoder, mapping the visual prompt to the same embedding space as the visual feature, which is then appended together with language token embeddings as the input of the LLM.
(3) \underline{Mask Decoder.} Instead of requiring an additional segmentation module for grounding language expressions onto 3D point clouds, OST directly generates 3D masks, keeping the model streamlined and self-contained.

Moreover, at the pretaining stage, OST is connected together with the 3D scene encoder and jointly pre-trained with the hybrid supervision of instance segmentation and 2D-to-3D knowledge distillation. Here, the 2D feature is extracted from multi-view images with the visual encoder of a 2D LMM, \emph{i.e.} LLaVA-1.5~\cite{liu2024improved}, and lifted to 3D by the geometric correspondence~\cite{peng2023openscene} between the point cloud and the pixels. 
Such a pretraining scheme on the one hand encompasses the perception prior to our model and takes the well-aligned 2D data as the bridge to facilitate the alignment between 3D visual embedding and language embedding.

We conduct end-to-end instruction tuning over various tasks and then benchmark our 3D-LLaVA on five popular 3D vision and language understanding datasets. 
As shown in Figure~\ref{fig:intro} Our method achieves the state-of-the-art performance on all of these datasets. Remarkably, we achieve 92.6\% CiDEr on the competitive ScanQA dataset, improving the previous best result by absolutely 4.9\% CiDEr score.

To summarize, we make three-fold contributions:
\begin{itemize} [leftmargin=2em]
    \item We propose 3D-LLaVA, a generalist 3D LMM that unifies multiple tasks through the Omni Superpoint Transformer, streamlining the framework.

    \item We present a new perspective that a versatile visual connector can be leveraged to remove the task-specific modules added to the 3D LMM, making the model more elegant and integrated.
    
    \item We benchmark the proposed method on different datasets, demonstrating its great potential to be a powerful baseline in this field.
    
\end{itemize}
\section{Related Work}
\label{sec:related_work}

\begin{figure*}
    \centering
    \includegraphics[width=0.96\linewidth]{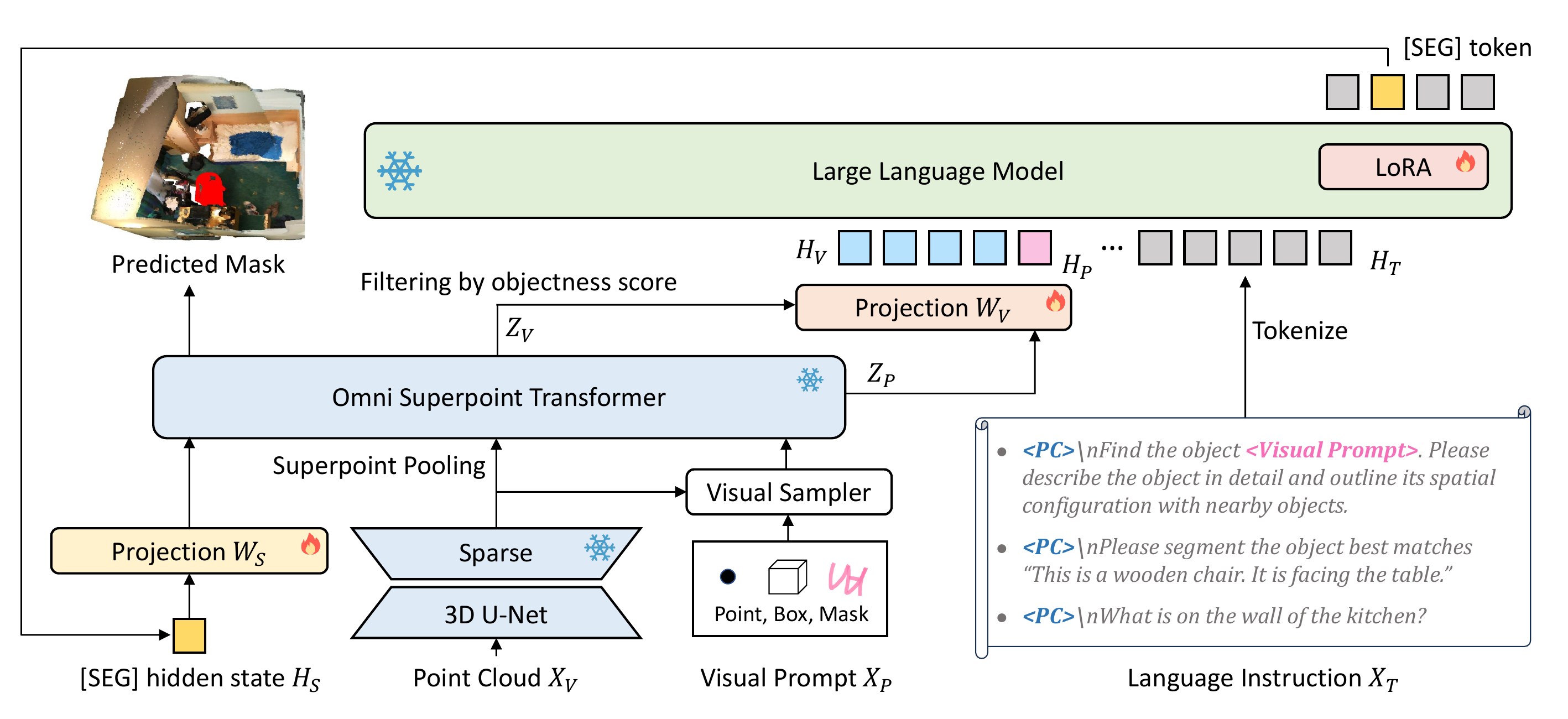}
    \vspace{-0.25cm}
    \caption{An overview of \sysname framework. Given input point cloud, language instruction, and optional visual prompt, 3D-LLaVA generates text output from LLM and produces 3D masks with Omni Superpoint Transformer (OST). The 3D feature out of the Sparse 3D U-Net is clustered into superpoint with Superpoint Pooling. Visual Sampler is a parameter-free module that samples point features corresponding to the visual prompt $X_P$. Omni Superpoint Transformer takes both superpoint feature and visual prompt feature as input, produces visual feature embedding $Z_V$ and visual prompt embedding $Z_P$, followed by a projection layer $W_V$ to obtain the token embedding $H_V$ and $H_P$. Once the LLM outputs a special segmentation token, \emph{i.e.}, [SEG], the hidden state linked to [SEG] token will be sent to another projection layer $W_S$ and then input as segmentation query to the frozen OST to generate segmentation masks.}
    \vspace{-0.2cm}
    \label{fig:framework}
\end{figure*}

\noindent \textbf{3D Vision \& Language Understanding.}
In recent years, there has been tremendous progress in understanding 3D scenes from natural language, where the language provides contextual knowledge and queries of user intentions to allow seamless interaction between humans and models. These works can be broadly categorized into four main tasks: 3D grounding~\citep{chen2020scanrefer,achlioptas2020referit3d,huang2022multi,zhang2023multi3drefer,yang2021sat} that localizes specific objects within the 3D scene according to the given textual queries, 3D referring segmentation~\citep{huang2021text,qian2024x,wu20243d,he2025segpoint,huang2024reason3d} that predicts a point-wise mask for the described object; 3D captioning~\citep{chen2021scan2cap,yuan2022x-trans2cap,jiao2022more,chen2023vote2cap-detr,chen2023vote2cap-detr++,jin2024tod3cap,kim2024bi} that densely localizes objects in a 3D scene and describes them with natural language; 3D question answering~\citep{azuma2022scanqa,parelli2023clip-guided,ma2022sqa3d} that answers given textual questions about the 3D scene. 

Although achieving great success on certain tasks, the above methods fall short in generalizing across different 3D understanding tasks. Motivated by this, recent efforts have also been dedicated to designing pre-training schemes~\citep{zhu20233d-vista,jin2023-3D-VLP} or unified models~\citep{cai20223djcg,chen2021d3net} for various tasks like 3D grounding, captioning and question answering. Despite these models achieving impressive improvements in handling diverse 3D scene tasks, their reliance on task-specific heads and limited reasoning capabilities constrain their flexibility for broader, general-purpose applications.

\noindent \textbf{3D Large Multimodal Model.}
The huge success of Large Language Models (LLMs)~\citep{brown2020language,touvron2023llama,chowdhery2023palm,hoffmann2022training} has fueled the demand for a versatile interface that can handle various modalities beyond language. In response to this demand, Large Multimodal Models (LMMs) has been developed to comprehend instructions that span vision and language~\citep{liu2023llava,alayrac2022flamingo,team2023gemini,zhang2024omg,liu2024improved}. PointLLM~\citep{xu2023pointllm} integrates the object-level point cloud into LLM by constructing a joint embedding space among 3D points and text, enabling explain the 3D shape with language. 3D-LLM~\citep{hong20233d-llm} extends the 2D LMM into the 3D scene, improving the capability of 3D spatial reasoning by introducing positional embeddings and location tokens. LL3DA~\cite{chen2024ll3da} develops a Q-Former~\cite{li2023blip2} to bridge the 3D point cloud, visual prompt, and language instruction. Grounded 3D-LLM~\cite{chen2024grounded} involves referent tokens and employs contrastive learning to unify grounding with textual response generation. Segpoint~\cite{he2025segpoint} attempts to unify semantic segmentation and referring segmentation with an LLM.  Agent3D-Zero~\cite{zhang2024agent3d} leverages the 2D LMM to first observe from the birds' eye view and then selects the informative viewpoints for further zero-shot 3D scene understanding. 
Scene-LLM~\cite{fu2024scene} lifts multi-view image features into 3D space, and follows the two-stage training scheme~\cite{liu2023llava} to to perform 3D vision and language alignment.
Chat-Scene~\citep{huang2024chat} proposes to achieve precise object referencing and grounding by incorporating object identifiers into the 3D LMM and fusing the offline extracted 2D and 3D instance-level features.
\section{Approach}

The overall framework of 3D-LLaVA is illustrated in Figure~\ref{fig:framework}. 
It is a generalist 3D LMM, capable of conducting 3D vision-centric dialogue, being interacting seamlessly with flexible visual and textual prompts, and grounding open-ended language description into 3D point cloud masks.
In this section, we first introduce the model architecture of the 3D scene encoder (Section~\ref{sec:encoder}) and Omni Superpooint Transformer (Section~\ref{sec:ost}). Then, in Section~\ref{sec:pipeline}, we elaborate on the detail of each step in our pipeline. Finally, in Section~\ref{sec:training}, we introduce the training scheme.

\subsection{3D Scene Encoder}\label{sec:encoder}
Given the point clouds input $\mathbf{X_V}\in\mathbb{R}^{N\times6}$, where N is the number of points and the 6 channels represent the coordinates $\{x, y, z\}$ and the color information $\{r, g, b\}$, we first convert the points into voxels based on their 3D coordinates. After obtaining the voxels, the Sparse 3D U-Net~\cite{choy20194d} is leveraged as the scene encoder to extract point cloud features. Sparse 3D U-Net is a U-Net-like architecture but consists of sparse convolutional layers. The output of the Sparse 3D U-Net has the same number as the input, resulting in an excessively large voxel count that is not feasible for the following steps. One option to reduce the number of points is to perform farthest point sampling~\cite{qi2017pointnet++}. However, the sampling operation inevitably causes information loss. In contrast, we follow ~\cite{sun2023superpoint,kolodiazhnyi2024oneformer3d,lai2023mask} to implement the average pooling operation based on superpoints, which are generated with the bottom-up clustering algorithm~\cite{landrieu2018large}. The superpoint pooling reduces the quantity of 3D vision embeddings into hundreds or a few thousand, depending on the complexity of the 3D scene.

\subsection{Omni Superpoint Transformer}\label{sec:ost}
\begin{figure}
    \centering
    \includegraphics[width=0.999\linewidth]{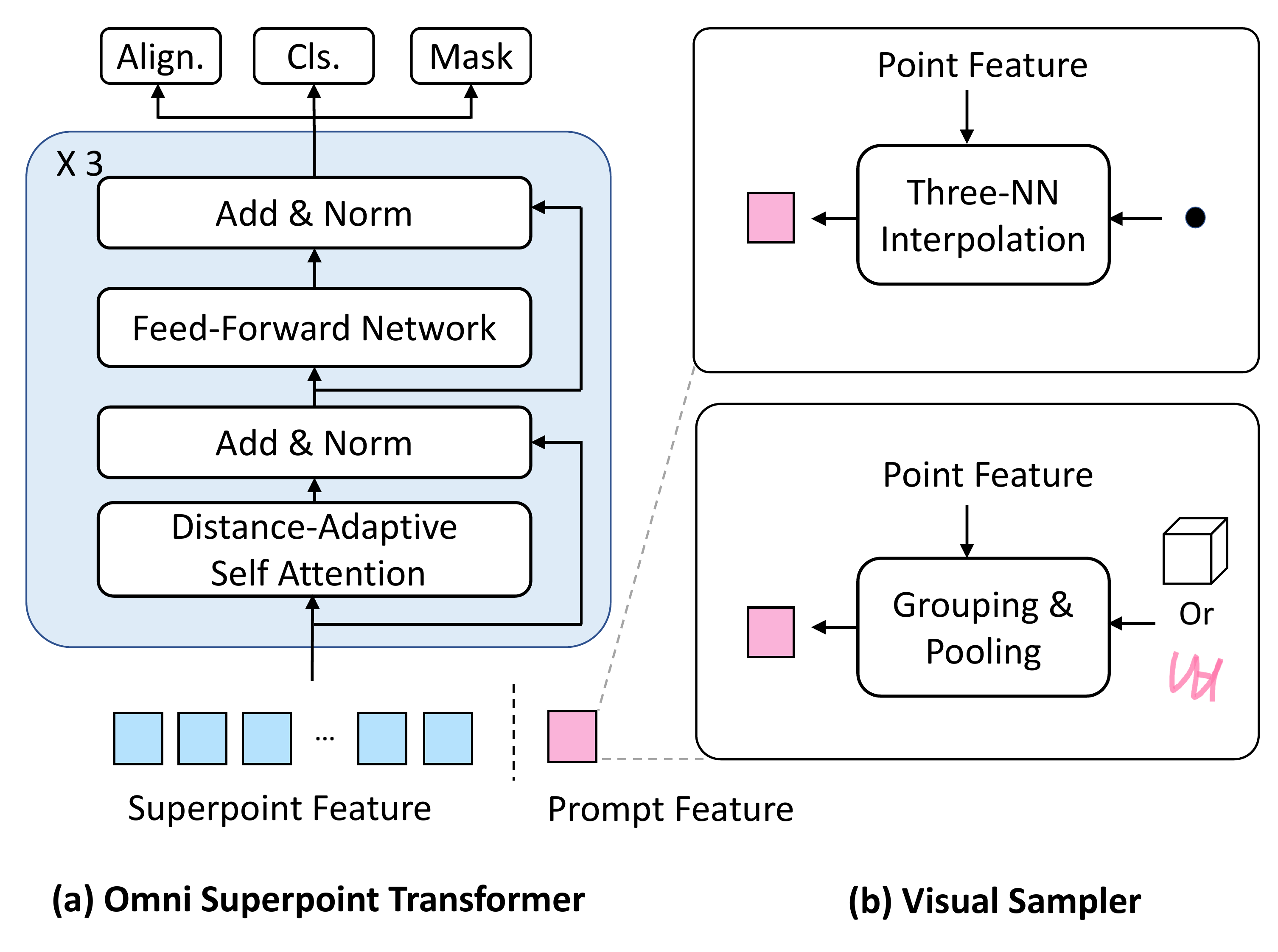}
    \caption{An illustration of (a) the architecture of Omni Superpoint Transformer, and (b) the paradigm of the visual sampler.}
    \label{fig:ost}
\end{figure}

The architecture of the proposed Omni Superpoint Transformer (OST) is shown in Figure~\ref{fig:ost} (a). Notably, the basic block of a conventional segmentation Transformer typically includes a cross-attention layer, a self-attention layer, and a feed-forward network. Here, the cross-attention layer is leveraged to abstract the information from the source feature to the object query.
Although OST can perform segmentation, it is primarily composed without cross-attention layers. The superpoint features act as both queries and source features (key and value) in OST. 
This adjustment keeps the correspondence between the output embedding of OST and the lifted 2D feature, facilitating effective 2D-to-3D feature distillation during the pretraining phase.
Additionally, to guide the superpoint queries towards relevant entities, we replace the standard self-attention layer with a distance-adaptive self-attention layer~\cite{liu2023sparsebev}, which introduces a bias term based on the distances between superpoints. The pairwise attention between the i-th superpoint query and the j-th superpoint query is computed as:
\begin{equation}
    Attn(Q_i, K_j, V_j) = Softmax(\frac{Q_iK_j^T}{\sqrt{C}} - \sigma \cdot D)V_j,
    \label{eq:1}
\end{equation}
where $Q, K, V$ is the query, key, and value of the attention module, $C$ is the channel of the embedding, $\sigma$ is a learnable parameter based on the query, and $D$ indicates the Euler distance between the centroid of these two superpoints.

There are three heads on the top of OST: a mask head, a classification head, and an alignment head. The mask head transforms each query embedding into a mask prediction kernel, which is then applied to generate binary mask prediction by performing a dot product with the input superpoint features of OST~\cite{sun2023superpoint,schult2023mask3d,jain2023oneformer}.
The classification head predicts the category of the segmentation mask by outputting the logit of each category.
The output of the alignment head is denoted as $Z_V$ in Figure~\ref{fig:framework}. It would be further leveraged to obtain the visual token of the LLM.

\subsection{Details in Pipeline}\label{sec:pipeline}

\noindent \textbf{Visual Feature Selection.} 
Although superpoint pooling has reduced the query number of OST, it still results in a very long sequence if directly applied as input visual tokens of the LLM. To alleviate this issue, after obtaining $Z_V$ from OST, we only keep the superpoints with the top-K objectness scores. The objections score of each superpoint query is defined as the highest score among foreground categories.

\noindent \textbf{Visual Prompt Encoding.} 
A generalist 3D LMM is supposed to be interacted with both language instructions and visual prompts. Common visual prompts include a clicking point, bounding box, or binary mask. A straightforward approach to encode these prompts is to use a prompt encoder composed of several linear layers~\cite{chen2024ll3da}, designed to project the prompt (e.g., coordinates of clicking points or bounding boxes) into an embedding that aligns with the same semantic space as visual or language tokens. However, we find that this type of prompt encoder is challenging to optimize, as it lacks explicit information indicating which areas are targeted by the prompt.

In contrast, as shown in Figure~\ref{fig:ost} (b), \sysname introduces a parameter-free visual sampler to encode the visual prompt $X_P$ and reuses OST as a visual prompt encoder to generate the corresponding visual prompt embedding $Z_P$, ensuring that the prompt is embedded in the same space as the visual features. 

For a clicking point prompt, the visual sampler obtains the prompt feature through three nearest-neighbor (three-NN) interpolation~\cite{qi2017pointnet++}, which first finds the three nearest points and computes the prompt feature using weighted interpolation. If the prompt is a box or mask, the visual sampler groups the points within the prompt and applies average pooling to generate the prompt feature. This prompt feature is then appended to the superpoint features and is input to OST. Here, we leverage the masked attention strategy between the superpoint feature and prompt feature. Specifically, we set the attention mask from the superpoints to the prompt as negative infinity. This prevents the prompt feature from influencing the superpoints, allowing it only to sample the relevant visual information.  Similar to the visual feature embedding $Z_V$, the prompt embedding $Z_P$ is out of the alignment head.

\noindent\textbf{Projection.}
After obtaining the Top-K superpoint-based visual feature embedding $Z_v$ and the visual prompt embedding $Z_P$, we apply a projection layer $W_V$ to transform them into the language embedding tokens $H_V$ and $H_P$. The projection layer consists of two linear layers and a GELU activation layer between the linear layers. 

\noindent\textbf{Instruction Formulation.}
We present the typical language instruction in the bottom-right part of Figure~\ref{fig:framework}. There are two kinds of place holders in the instruction: ``$\textlangle$PC$\textrangle$'' and ``$\textlangle$Visual Prompt$\textrangle$''. The text instruction except for the placeholder will be tokenized into the text token embedding $H_T$. After tokenization, we replace ``$\textlangle$PC$\textrangle$'' with visual token embedding $H_V$, and replace ``$\textlangle$Visual Prompt$\textrangle$'' with the prompt token embedding $H_P$.

\noindent\textbf{Mask Decoding.}
When the instruction prompts 3D-LLaVA to perform referring segmentation~\cite{chen2020scanrefer,achlioptas2020referit3d,lai2024lisa}, the LLM will output a [SEG] token in its text response. 
Once detecting this token, we extract the last hidden state of the token before the [SEG] token. This hidden state, $H_S$, is then fed into the projection layer $W_S$ to generate the segmentation query. 

In our method, we leverage the frozen OST to predict the segmentation mask of the referred object. Similar to the paradigm of using OST as the visual prompt encoder, the segmentation query is concatenated with the superpoint query to formulate the input of OST. We apply a mask attention strategy to prevent information flow from the segmentation query to the superpoints. Since the segmentation query lacks coordinate information, the bias term (from Equation~\ref{eq:1}) between this query and the superpoint queries is set to zero.
The output kernel from the mask head that corresponds to the segmentation query is applied to the superpoint feature input to generate the mask prediction.

\subsection{Training Scheme} \label{sec:training}
\noindent \textbf{Stage 1: Pre-training 3D Scene Encoder and OST.} Unlike the 2D domain, which has powerful and widely recognized vision foundation models such as CLIP~\cite{radford2021learning}, there is currently no such 3D foundation model that can serve as a readily usable 3D visual encoder.

To this end, we pre-train the Sparse 3D U-Net and the OST by ourselves. Specifically, we adopt hybrid supervision that combines the vision-centric task, \emph{i.e.}, instance segmentation, and the 2D-to-3D  knowledge distillation:
\begin{equation}
    L_{Pre} = L_{Cls} + L_{Mask} + L_{KD},
\end{equation}
where $L_{Cls}$ represents cross-entropy loss for multi-category classification, $L_{Mask}$ includes the binary cross-entropy loss and Dice loss for mask prediction, and $L_{KD}$ denotes the knowledge distillation loss, which includes mean squared error and cosine similarity losses.

For instance segmentation, we leverage the annotation from ScanNet200~\cite{rozenberszki2022language} as the training data. For 2D-to-3D knowledge distillation, we follow OpenScene~\cite{peng2023openscene} that first extracts multi-view 2D features and then lifts the 2D features into 3D points by the correspondence between 3D point clouds and 2D pixels. The lifted 2D features are pooled into each superpoint to generate the target feature. Here we leverage the visual encoder of LLaVA-1.5-7B~\cite{liu2024improved}, \emph{i.e.,} CLIP-ViT-L, to extract the teacher 2D feature.

\begin{table}[t]
    \centering
    \setlength{\tabcolsep}{12pt}
    \small
    \caption{\textbf{Dataset statistics for joint instruction tuning.}}
    \vspace{-0.1cm}
    \renewcommand{\arraystretch}{1.1}

    \begin{tabular}{l|c|r}
        \hline
        \hline
\rowcolor[gray]{.92}        Dataset & Task & Size \\
        \hline
         ScanRefer & referring segmentation &37K \\
         Nr3D & referring segmentation &29K \\
         Multi3DRefer &referring segmentation & 44K \\
         ScanQA & visual question answering &30K \\
         SQA3D & visual question answering &89K \\
         Scan2Cap & dense captioning &37K \\
         Nr3D$^*$ & dense captioning & 29K \\
        \hline
 \rowcolor{tabblue!10} Total & - & 295K\\
        \hline
        \hline
    \end{tabular}
    \label{tab:dataset_statistics}
    \vspace{-0.2cm}
\end{table}

\noindent \textbf{Stage 2: End-to-End Instruction Tuning.} We combine various 3D vision and language understanding datasets to form our instruction-tuning data. The combined datasets include ScanRefer~\cite{chen2020scanrefer}, Nr3D~\cite{achlioptas2020referit3d}, Multi3DRefer~\cite{zhang2023multi3drefer}, ScanQA~\cite{azuma2022scanqa}, SQA3D~\cite{ma2022sqa3d}, Scan2Cap~\cite{chen2021scan2cap}. The statistic of the utilized dataset is presented in Table~\ref{tab:dataset_statistics}. To enrich the language annotation that involves describing the object in the 3D scene, we follow~\cite{huang2024chat} to use Nr3D as the complementary to the dense captioning task, which is denoted as ``Nr3D*'' in this table.

The instruction tuning phase jointly optimizes 3D-LLaVA for both text generation and referring segmentation. The training objective is composed as follows:
\begin{equation}
    L_{IFT} = L_{text} + 0.1 \times   L_{mask},
\end{equation}
where $L_{text}$ is the cross-entropy loss for next-token generation, $L_{mask}$ represents the mask prediction loss that also consists of the binary cross-entropy loss and the Dice loss, which is the same as the pertaining stage. Here, we multiply the mask loss by a coefficient of 0.1 for balance.
We always keep the Sparse 3D U-Net, OST, and the main body of LLM frozen. Only the visual projector, the SEG project, and LoRA~\cite{hu2021lora} parameters adopted to the LLM are updated.

\section{Experiments}

\begin{table*}[htbp]
    \caption{
        \textbf{Performance comparison among state-of-the-art methods.} ``Specialist Model'' means this model can be utilized to perform 3D question answering, 3D dense captioning, or referring segmentation. ``Finetuned 3D LMM'' indicates the model is jointly trained and then finetuned on each dataset before evaluation. We add a ``*'' to  3D LMMs that belong to this kind. ``3D LMM'' includes the models that are only been trained on multiple tasks. ``PC'' means point cloud and ``I'' means multi-view images. Please note that LEO~\cite{huang2023embodied}'s results on ScanQA is marked with a gray color and not compared to other methods, since it is in a different setting that accesses the ground truth object related to the question. The top-2 entities of each metric are marked with \underline{underline} and the best one is highlighted by bolding font.
    }
    \vspace{-0.2cm}
    \renewcommand{\arraystretch}{1.05}
    \centering
    \resizebox{0.99\linewidth}{!}{
    \begin{tabular}{l|c|c|c|cccc|cc|cccc}
    \hline
    \hline
    \rowcolor[gray]{.92} &  & ScanRefer (val) & Multi3DRefer (val) & \multicolumn{4}{c|}{ScanQA (val)} & \multicolumn{2}{c|}{SQA3D (test)} & \multicolumn{4}{c}{Scan2Cap (val)} \\
     \rowcolor[gray]{.92} Method & Modality & mIoU$\uparrow$ & mIoU$\uparrow$ & C$\uparrow$ & B-4$\uparrow$ & M$\uparrow$ & R$\uparrow$ & EM$\uparrow$ & EM-R$\uparrow$ & C@0.5$\uparrow$ & B-4@0.5$\uparrow$ & M@0.5$\uparrow$ & R@0.5$\uparrow$ \\ \hline
    \textbf{\textit{Specialist Models:}} & & & & & & & & & & & & & \\
    ScanQA\cite{azuma2022scanqa}   & PC & - & - & 64.9 & 10.1 & 13.1 & 33.3 & 46.6 & - & - & - & - & - \\
    3D-VLP\cite{jin2023-3D-VLP} & PC & - & - & 67.0 & 11.2 & 13.5 & 34.5 & - & - & 54.9 & 32.3 & 24.8 & 51.5 \\
    3D-VisTA\cite{zhu20233d-vista} & PC & - & - & 69.6 & 10.4 & 13.9 & \underline{\textbf{45.7}} & 48.5 & - & 61.6 & 34.1 & 26.8 & 55.0 \\
    Scan2Cap\cite{chen2021scan2cap} & PC & - & - & - & - & - & - & 41.0 & - & 39.1 & 23.3 & 22.0 & 44.8 \\
    MORE\cite{jiao2022more} & PC & - & - & - & - & - & - & - & - & 40.9 & 22.9 & 21.7 & 44.4 \\
    SpaCap3D\cite{wang2022spacap3d} & PC & - & - & - & - & - & - & - & - & 44.0 & 25.3 & 22.3 & 45.4 \\
    D3Net\cite{chen2021d3net} & PC & - & - & - & - & - & - & - & - & 46.1 & 30.3 & 24.4 & 51.7 \\
    UniT3D\cite{chen2023unit3d} & PC & - & - & - & - & - & - & - & - & 46.7 & 27.2 & 21.9 & 46.0 \\
    3DJCG\cite{cai20223djcg} & PC & - & - & - & - & - & - & - & - & 49.5 & 31.0 & 24.2 & 50.8 \\
    Vote2Cap-DETR~\cite{chen2023vote2cap-detr} & PC & - & - & - & - & - & - & - & - & 61.8 & 34.5 & 26.2 & 54.4 \\
    TGNN~\cite{huang2021text} & PC & 27.8 & -& -& - & - & - & - & - & - & - & - & - \\
    M3DRef-CLIP~\cite{zhang2023multi3drefer} & PC & 35.7 & 32.6& -& - & - & - & - & - & - & - & - & - \\
    X-RefSeg3D~\cite{qian2024x} & PC & 29.9 & - & -& - & - & - & - & - & - & - & - & - \\
    
    3D-STMN~\cite{wu20243d} & PC & 39.5 & - & -& - & - & - & - & - & - & - & - & - \\
    \hline
    \textbf{\textit{Finetuned 3D LMMs:}} & & & & & & & & & & & & & \\
    3D-LLM\cite{hong20233d-llm} & PC+I & - & - & 69.4 & 12.0 & 14.5 & 35.7 & - & - & - & - & - & - \\
    Scene-LLM~\cite{fu2024scene}$^*$ & PC+I & - & - & 80.0 & 12.0 & 16.8 & 40.0 & 54.2 & - & - & - & - & - \\
    LL3DA$^*$~\cite{chen2024ll3da} & PC & - & - & 76.8 & 13.5 & 15.9 & 37.3 & - & - & 65.2 & 36.8 & 26.0 & 55.1 \\
    SegPoint~\cite{he2025segpoint}$^*$ & PC & \underline{41.7} & \underline{36.1} & - & - & - & - & - & - & - & - & - & - \\
    \hline
    \textbf{\textit{3D LMMs:}} & & & & & & & & & & & & & \\
    LEO~\cite{huang2023embodied} & PC+I &- & - & \textcolor{gray}{101.4} & \textcolor{gray}{13.2} & \textcolor{gray}{20.0} & \textcolor{gray}{49.2} & 50.0 & 52.4 & 72.4& \underline{\textbf{38.2}} & \underline{\textbf{27.9}} &  \underline{\textbf{58.1}} \\
    Scene-LLM~\cite{fu2024scene} & PC+I & - & - & 80.0 & 11.7 & 15.8 & 35.9 & 53.6 & - & - & - & - & - \\
    Chat-Scene~\cite{huang2024chat} & PC+I & - & - & \underline{87.7} & \underline{14.3} & \underline{18.0 }& 41.6 & \underline{\textbf{54.6}} & \underline{\textbf{57.5}} & \underline{77.2} & 36.4 & \underline{\textbf{28.0}} & \underline{\textbf{58.1}}\\
    Grounded 3D-LLM~\cite{chen2024grounded} & PC & - & - & 72.7 & 13.4 & - & - & - & - & 70.6 & 35.5 & - & - \\
    \rowcolor{tabblue!10}  3D-LLaVA (ours) & PC & \underline{\textbf{43.3}} & \underline{\textbf{42.7}} & \underline{\textbf{92.6}} & \underline{\textbf{17.1}} & \underline{\textbf{18.4}} & \underline{43.1} & \underline{54.5} & \underline{56.6} & \underline{\textbf{78.8}} & \underline{36.9} & 27.1 & 57.7 \\
    \hline
    \hline
    \end{tabular}
    }
    \label{tab:benchmark-all}
    \vspace{-0.3cm}
\end{table*}

\subsection{Datasets and Metrics}
\textbf{Datasets.}
In this work, we conduct experiments on the 3D scans provided by ScanNet dataset~\cite{dai2017scannet}, including 1,201 scenes for training and 312 for validation. At the pertaining stage of our 3D encoder, we leverage the mask annotation from ScanNet200~\cite{rozenberszki2022language}, which extends the original ScanNet with fine-grained categories. The language annotation leveraged in the instruction tuning has been introduced in Section 4. After instruction tuning, 
we validate the effectiveness of the proposed \sysname on the following datasets: ScanQA~\cite{azuma2022scanqa}  and SQA3D~\cite{ma2022sqa3d} for question answering, ScanRefer~\cite{chen2020scanrefer} and Multi3DRefer~\cite{zhang2023multi3drefer} for referring segmentation and Scan2Cap~\cite{chen2021scan2cap} for dense captioning.

\noindent \textbf{Metrics.}
We follow the common practice to evaluate the quality of generated text response for ScanQA and Scan2Cap in terms of CiDEr (\textbf{C}) BLEU-4 (\textbf{B-4}), METEOR (\textbf{M}) and Rouge-L (\textbf{R}). 
Different from the conventional setting of ScanQA, there is a definite answer to situated question answering dataset SQA3D, therefore we leverage extract match accuracy (\textbf{EM}) as well as the refined version (\textbf{EM-R}) as the metric. For referring segmentation, we adopt the mean intersection over union (\textbf{mIoU}) for evaluation.

\subsection{Implementation Details} 
We pre-train our 3D visual encoder on ScanNet200 for 512 epochs under the hybrid supervision of 2D-to-3D knowledge distillation and segmentation. After obtaining the 3D visual encoder, we developed our \sysname based on the LLaVA-1.5-7B~\cite{liu2024improved}. We make use of model weights of the visual projector and LLM (Vicuna-1.5-7B~\cite{vicuna2023}) from the LLaVA-1.5-7B, and connect the alignment embedding out of our 3D visual encoder to the visual projector.
We keep 100 superpoint features $Z_V$ according to their objectness score, which are then been projected to the visual token embeddings $H_V$.
The instruction tuning is conducted on 8$\times$ NVIDIA RTX 3090 GPUs with the acceleration of the DeepSpeed toolkit. We adopt LoRA~\cite{hu2021lora} to the LLM and keep the main body of LLM and visual encoder frozen during training. The data presented in Table~\ref{tab:dataset_statistics} are leveraged to perform end-to-end training for 1 epoch. We set the batch size to 2 for each GPU and update the model weights after accumulating the gradient every 8 steps. The model is optimized with the AdamW. The Cosine Annealing schedule is leveraged to update the learning rate, with the initial learning rate set as 2e-4.

\subsection{Comparison with SoTA Models}
We compare the proposed 3D-LLaVA with other models and present the results in Table~\ref{tab:benchmark-all}. The models compared in this table are divided into three groups: specialist models, Finetuned 3D LMMs, and 3D LMMs. The specialist model is designed to address a single kind of task. All of the specialist models in this table are without LLMs. The Finetuned 3D LMM is the 3D large multimodal model that is finetuned on each dataset. Such fine-tuning could improve the performance of the model on the corresponding dataset, but will affect its generalizability. The last kind, 3D LMM, is the large multimodal model that is trained on a unified dataset including various tasks. Particularly, among all the competitors, our 3D LLaVA is the only one that covers the typical text generation task (\emph{i.e.}, 3D dense captioning, and 3D vision question answering) and point-level understanding (\emph{i.e.}, 3D Referring Segmentation).

\noindent \textbf{3D Referring Segmentation} requires the model to output the 3D mask on the point cloud according to the user's language expression, which validates the capability of grounding the text description on the 3D scene. 
We benchmark our methods and other state-of-the-art methods on both the single-target setting (Scanerfer~\cite{chen2020scanrefer}) and the various number setting (Multi3DRefer~\cite{zhang2023multi3drefer}).The referring text from Multi3DRefer can correspond to one, many, or even zero objects. If multiple objects are referred to in the instruction, we follow~\cite{he2025segpoint} to merge the masks into a single one for evaluation. When there is no corresponding to the referring expression, our 3D-LLaVA will output ``Sorry, I cannot find this object.''. In this case, since there is no [SEG] token in the response, the mask decoding pipeline will not be applied, and thus all of the predicted masks will be assigned as background. As shown in the table, our 3D-LLaVA reports the best result among the competitors.
Notably, our model achieves 43.3\% mIoU on ScanRefer and 42.7\% mIoU on Multi3DRefer, improving the previous best record of SegPoint by 1.6\% mIoU and 6.6\% mIoU.

\begin{figure}
    \centering
    \includegraphics[width=0.99\linewidth]{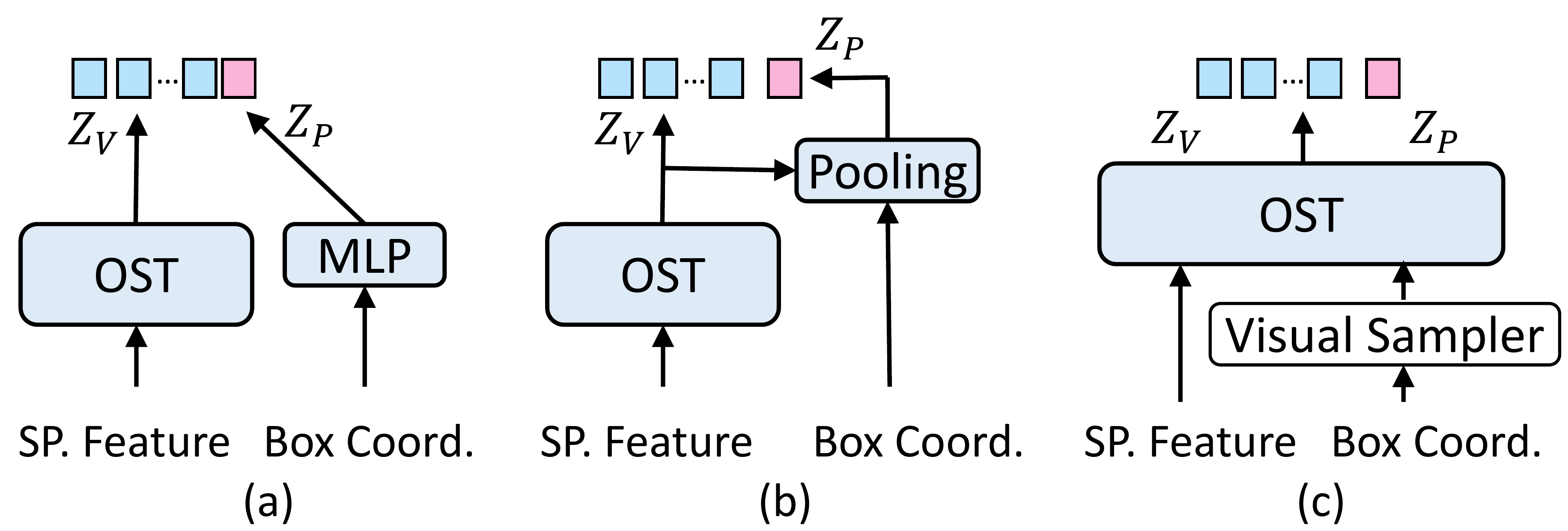}
    \vspace{-0.15cm}
    \caption{Different paradigms to produce visual prompt embedding. ``OST'': Omni Superpoint Transformer. ``P.E. Encoder'': Parameter-Free Encoder. }
    \label{fig:compare_prompt}
    \vspace{-0.5cm}
\end{figure}

\noindent \textbf{3D Question Answering} is the task that asks the model to observe the visual information of the 3D scene and give a precise response to the user's question involving some part of the scene. 
We conduct the comparison between our 3D-LLaVA and other methods on both the conventional 3D question-answering dataset ScanQA~\cite{azuma2022scanqa} and the situated question-answering dataset SQA3D~\cite{ma2022sqa3d}. As shown in Table~\ref{tab:benchmark-all}, our method ranks the best for CiDEr, BLEU-4, and METEOR among the methods without accessing ground-truth information of the object relevant to the question. Remarkably, compared to Grounded 3D-LLM~\cite{chen2024grounded} which only uses point cloud as input, our 3D-LLaVA achieves a 19.9\% improvement. When compared to the strongest competitor Chat-Scene, our 3D-LLaVA achieves 4.9\% CiDEr, 2.8\% BLEU-4, 0.4\% METEOR, and 1.4\% Rouge-L improvements on ScanQA, respectively. On SQA3D, our 3D-LLaVA reports comparable extract match accuracy as that of Chat-Scene (54.5\% V.S. 54.6\%). It is worth noting that Chat-Scene uses both instance-level 3D and 2D features, which rely on complicated offline preprocessing, while our 3D-LLaVA extracts superpoint features online with the OST, which is more computation-friendly.

\noindent \textbf{3D Dense Captioning} demands the model to describe the object and its spatial relationship to the surrounding instances within the scene. In this experiment, we follow the common practice of using the predicted mask proposals of Mask3D~\cite{schult2023mask3d} as the visual prompt. Please note that we have not got access to the output of Mask3D in the training stage. Our OST works as a visual sampler to convert any prompts in the predefined formulation to the semantic space of visual features without the extra cost of finetuning. Results in the table show that our 3D-LLaVA also achieves the best performance in generating instance-level descriptions. This experiment further validates the effectiveness and scalability of the proposed 3D-LLaVA with OST.

\begin{table}[t]
    \centering
    \setlength{\tabcolsep}{12.0pt}
    \small
    \caption{\textbf{Performance comparisons for box-level 3D visual grounding.} Our results are obtained by directly converting the foreground referring mask to a box, highlighted with \colorbox{tabblue!10}{light blue.}
    }
    \vspace{-0.1cm}
    \renewcommand{\arraystretch}{1.2}

    \begin{tabular}{l|cc}
        \hline
        \hline
        \rowcolor[gray]{.92}   & \multicolumn{2}{c}{ScanRefer (Box-Level)} \\
        \rowcolor[gray]{.92} Methods & Acc@0.25 &Acc@0.5 \\
        \hline
        \textbf{\textit{Specialist Models:}} & & \\
        3D-VisTA~\cite{zhu20233d-vista} & 50.6 & 45.8 \\
        ConcretNet~\cite{unal2024four} & 50.6 & 46.5 \\
        \hline
        \textbf{\textit{3D LMMs:}} & & \\
        3D-LLM~\cite{hong20233d-llm} & 30.3 & - \\
        Grounded 3D-LLM~\cite{chen2024grounded} & 48.6 & 44.0 \\
        Chat-Scene~\cite{huang2024chat} & 55.5 & 50.2 \\
         \rowcolor{tabblue!10} 3D-LLaVA (Ours) & 51.2 & 40.6 \\
        \hline
        \hline
    \end{tabular}
    \label{tab:box_level_vg}
\end{table}

\begin{table}[t]
    \centering
    \setlength{\tabcolsep}{6.8pt}
    \small
    \caption{\textbf{Ablation study on the paradigm to produce visual prompt embedding.} The models are compared in terms of CiDEr, BLEU-4, METEOR and Rough-L on Scan2Cap~\cite{chen2021scan2cap}. The index (a), (b), and (c) in this table correspond to paradigms depicted in Figure~\ref{fig:compare_prompt}. Our default setting is highlighted with \colorbox{tabblue!10}{light blue.}}
    \vspace{-0.1cm}
    \renewcommand{\arraystretch}{1.2}

    \begin{tabular}{l|cccc}
        \hline
        \hline
        \rowcolor[gray]{.92}  & \multicolumn{4}{c}{Scan2Cap} \\
        \rowcolor[gray]{.92} Visual Prompt Encoding & C$\uparrow$ & B-4$\uparrow$ & M$\uparrow$ & R$\uparrow$ \\
        \hline
         (a) Coordinate Projection & 68.7 & 33.9 & 26.7 &  55.1\\
         (b) Pooling & 76.8 & 36.6 & 26.9 & 57.5 \\
           \rowcolor{tabblue!10} (c) Ours with OST & 78.8 & 36.9 & 27.1 &  57.7 \\
        \hline
        \hline
    \end{tabular}
    \label{tab:prompt_encoding}
\end{table}

\subsection{Experimental Analysis}
This section presents the experimental analysis of our 3D-LLaVA. Unless specified, the model evaluated in this section is trained with the same data and training scheme as the default setting introduced in the former sections.

\noindent\textbf{Evaluating Masks with the Box-level Metric.} Even not designed for 3D referring segmentation, our 3D-LLaVA can also produce box-level grounding results. Specifically, we first apply DBSCAN algorithm~\cite{ester1996density} to the foreground mask to remove outliers, and then
obtain the grounding box by considering the minimum and maximum coordinates of the mask. We compare the box-level grounding performance of our 3D-LLaVA with bot the specialist model and other 3D LMMs in Table~\ref{tab:box_level_vg}. Although our model is optimized for precise binary masks, whereas competitors are trained to select best-matching proposals based on box IoU, our 3D-LLaVA achieves 51.2\% accuracy when the IoU threshold is 0.25, better than most of the competitors in the table. Our performance lags behind Chat-Scene~\cite{huang2024chat}, but our method relies on neither an extra mask proposal generator nor the fusion of image and point cloud features. 

\noindent\textbf{Effect of Visual Prompt Encoding.} In this study, we analyze the effect of different ways to convert visual prompts into prompt embeddings (as illustrated in Figure~\ref{fig:compare_prompt}). We leverage the box as the visual prompt in this experiment since the mask can be converted to a box by its boundary and the clicking point is a special case of a box without area. Among the compared paradigms, our proposed strategy to reuse OST as the visual prompt encoder, \emph{i.e.} method (c), achieves the best result. On the one hand, our method avoids additional learnable parameters, which are difficult to optimize together with the LLM. On the other hand, compared to (b), appending the prompt query out of the parameter-free encoder to the superpoint queries enables deeply abstracting the superpoint features by the stack of OST encoder layers. 
The method (a) produces prompt embedding by applying an MLP to the box coordinates. This is because the produced prompt embedding lacks visual context, increasing the burden on the LLM in locating the corresponding region. We suppose this kind of paradigm needs more training data and training epochs to converge.

\begin{table}[t]
    \centering
    \setlength{\tabcolsep}{11pt}
    \small
    \caption{\textbf{Quantitative comparison on the different number of visual tokens.} The models are compared in terms of CiDEr and BLEU-4 on ScanQA~\cite{azuma2022scanqa} and Scan2Cap~\cite{chen2021scan2cap}. Our default setting is highlighted with \colorbox{tabblue!10}{light blue.}}
    \renewcommand{\arraystretch}{1.2}

    \begin{tabular}{l|cc|cc}
        \hline
        \hline
        \rowcolor[gray]{.92}  & \multicolumn{2}{c|}{ScanQA} & \multicolumn{2}{c}{Scan2Cap} \\
        \rowcolor[gray]{.92} \# Visual Token & C$\uparrow$ & B-4$\uparrow$ & C$\uparrow$ & B-4$\uparrow$ \\
        \hline
         50 & 91.1 & 15.9 & 74.9 & 35.9\\
         \rowcolor{tabblue!10} 100 & 92.6& 17.1 & 78.8 & 36.9 \\
         200 &92.8  &17.1 & 78.6&  37.2\\
         400 &92.3 &16.9 & 77.7&36.8 \\
        \hline
        \hline
    \end{tabular}
    \vspace{-0.2cm}
    \label{tab:token_number}
\end{table}

\noindent\textbf{Effect of Visual Token Number.} 
Retaining more visual tokens leads to a rapid increase in the computation complexity of LLMs. We take this experiment to explore how many visual tokens should be exploited in our 3D-LLaVA to enable an accurate understanding of the 3D scene. As shown in Table~\ref{tab:token_number}, increasing the token number from 50 to 100, the CiDEr on ScanQA and Scan2Cap is improved by 1.5\% and 3.9\%, respectively.  However, further increasing the token count to 200 yields no substantial performance gains. Therefore, we set the default token number to 100.

\begin{figure}[t]
    \centering
    \includegraphics[width=0.98\linewidth]{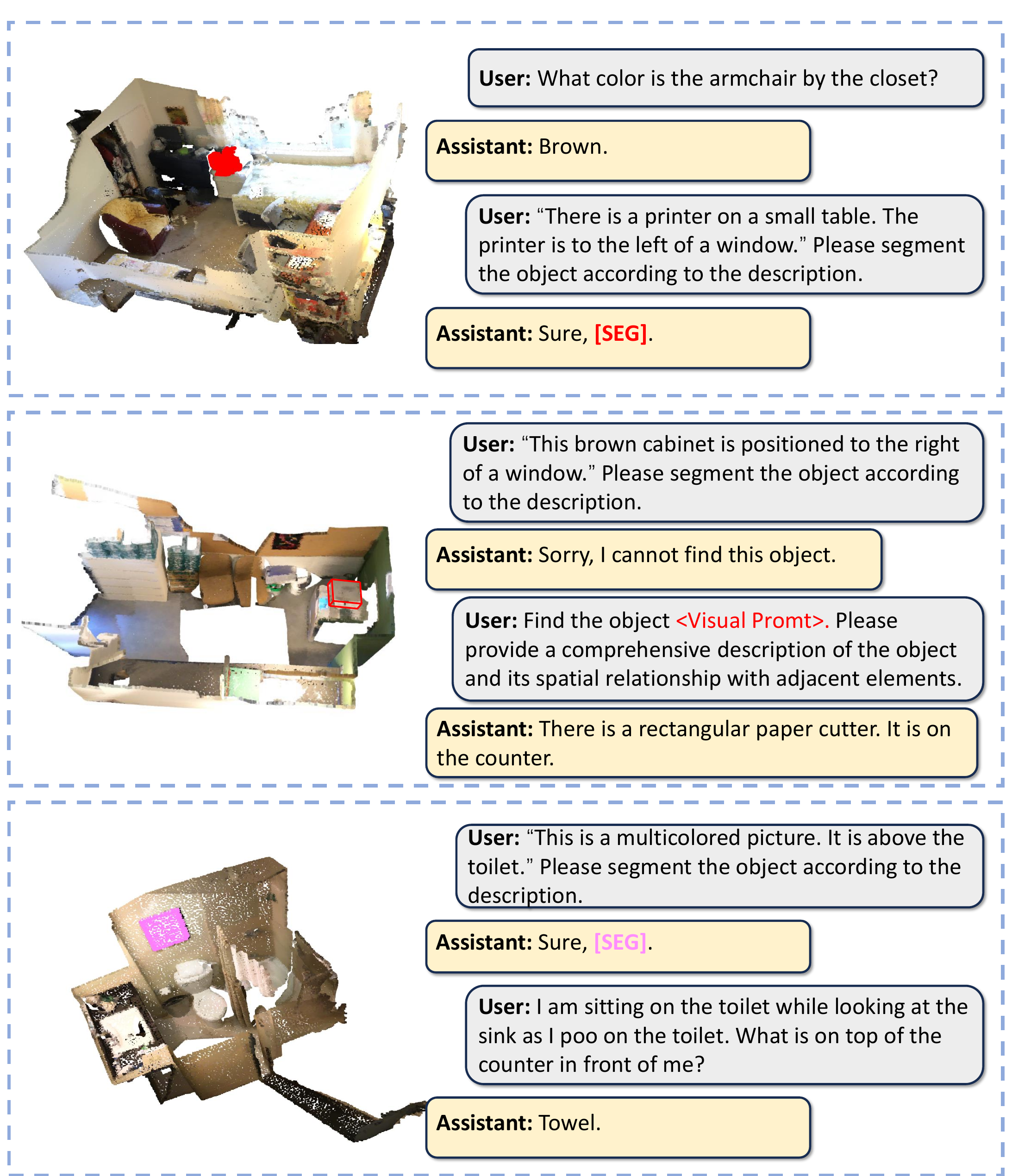}
    \caption{Visualization of 3D-LLaVA's response on various tasks. Each of these examples includes an instruction to perform referring segmentation. Besides, the examples present the result of 3D question answering~\cite{azuma2022scanqa}, 3D dense captioning~\cite{zhong2022contextual3DdenseCap}, and situated question answering~\cite{ma2022sqa3d}, respectively. When the referred object is not in the given 3D scene, the model is aware of responding with ``Sorry, I cannot find this object''.}
    \label{fig:visualization}
\end{figure}

\subsection{Qualitative Results}
In Figure~\ref{fig:visualization}, we showcase several visualizations of 3D-LLaVA's performance across various 3D environments, including bedrooms, offices, and bathrooms. Our 3D-LLaVA model accurately interprets user instructions and demonstrates an ability to avoid false positives when the target object is absent from the 3D scene.

\section{Conclusion}

In this work, we introduce 3D-LLaVA, a new 3D LMM with streamlined architecture and powerful capability. The core component in 3D-LLaVA is a new visual connector, Omni Superpoint Transformer (OST), which serves as a multifunctional module in visual token selection, visual prompt encoding, and mask decoding. Therefore, taking advantage of the versatile OST, 3D-LLaVA is capable of conducting 3D vision-centric dialogue, enabling flexible interaction and grounding language expression into 3D point cloud masks with a universal architecture. Through extensive experiments, 3D-LLaVA achieves impressive results across multiple benchmarks. Although 3D-LLaVA has made significant improvements over the previous methods, 3D data is still the main obstacle in developing 3D LMMs. We regard the data collection and configuration as the next step.

\noindent\textbf{Acknowledgements:} This work was supported by the Centre for Augmented Reasoning, an initiative by the Department of Education, Australian Government.

{
    \small
    \bibliographystyle{ieeenat_fullname}
    \bibliography{main}
}

\end{document}